\documentclass[conference]{IEEEtran}
\IEEEoverridecommandlockouts
\usepackage{cite}
\usepackage{amsmath,amssymb,amsfonts}
\usepackage{algorithmic}
\usepackage{graphicx}
\usepackage{textcomp}
\usepackage{xcolor}
\def\BibTeX{{\rm B\kern-.05em{\sc i\kern-.025em b}\kern-.08em
    T\kern-.1667em\lower.7ex\hbox{E}\kern-.125emX}}
    
\begin{document}

\title{From Data to Decision: Data-Centric Infrastructure for Reproducible ML in Collaborative eScience\\

\thanks{This work was supported by the Donaghue Foundation}
}

\author{\IEEEauthorblockN{Zhiwei Li}
\IEEEauthorblockA{\textit{Dept. of Industrial and Systems Engineering} \\
\textit{University of Southern California}\\
Los Angeles, USA \\
0009-0003-2848-7711}
\and
\IEEEauthorblockN{Carl Kesselman}
\IEEEauthorblockA{\textit{Information Sciences Institute} \\
\textit{University of Southern California}\\
Marina del Rey, USA \\
0000-0003-0917-1562}
\and
\IEEEauthorblockN{Tran Huy Nguyen}
\IEEEauthorblockA{\textit{Dept. of Computer Science} \\
\textit{University of Southern California}\\
Los Angeles, USA \\
0009-0008-2151-1025}
\and
\IEEEauthorblockN{Benjamin Yixing Xu}
\IEEEauthorblockA{\textit{Dept. of Ophthalmology} \\
\textit{University of Southern California}\\
Los Angeles, USA \\
0000-0003-1573-988X}
\and
\IEEEauthorblockN{Kyle Bolo}
\IEEEauthorblockA{\textit{Dept. of Ophthalmology} \\
\textit{University of Southern California}\\
Los Angeles, USA \\
0000-0001-7885-192X}
\and
\IEEEauthorblockN{Kimberley Yu}
\IEEEauthorblockA{\textit{Dept. of Ophthalmology} \\
\textit{University of Southern California}\\
Los Angeles, USA \\
0000-0002-3993-923X}
}

\maketitle

\begin{abstract}
Reproducibility remains a central challenge in machine learning (ML), especially in collaborative eScience projects where teams iterate over data, features, and models. 
Current ML workflows are often dynamic yet fragmented, relying on informal data sharing, ad hoc scripts, and loosely connected tools. This fragmentation impedes transparency, reproducibility, and the adaptability of experiments over time.
This paper introduces a data-centric framework for lifecycle-aware reproducibility, centered around six structured artifacts: Dataset, Feature, Workflow, Execution, Asset, and Controlled Vocabulary. These artifacts formalize the relationships between data, code, and decisions, enabling ML experiments to be versioned, interpretable, and traceable over time. The approach is demonstrated through a clinical ML use case of glaucoma detection, illustrating how the system supports iterative exploration, improves reproducibility, and preserves the provenance of collaborative decisions across the ML lifecycle.
\end{abstract}


\section{Introduction}
As machine learning (ML) becomes increasingly central to scientific discovery, concerns about correctness and reproducibility have grown~\cite{kapoor2023leakage}. In eScience, ML development is typically a collaborative and iterative process involving domain experts, data engineers, and ML researchers. These teams refine models based on evolving hypotheses and new data, creating feedback loops across data curation, feature engineering, modeling, and evaluation~\cite{10254870}. This dynamic process frequently introduces data cascades, where early curation errors propagate downstream, compounding over time~\cite{10.1145/3411764.3445518}.
In practice, ML workflows remain fragmented: datasets are shared informally, experiments span personal and cloud environments, and data, code, and configurations are often loosely coupled~\cite{hanson2023garbage}. These disjointed practices limit traceability and hinder reproducibility, creating significant risks in high-stakes domains like healthcare and genomics.
While MLOps and data management tools address parts of this problem, such as code versioning, pipeline orchestration, or environment encapsulation, they often overlook the full scientific lifecycle and the socio-technical realities of collaborative ML projects~\cite{kreuzberger2023machine}. 

In prior work, we introduced Deriva-ML~\cite{li2024deriva}, a socio-technical platform that extends the FAIR principles (Findable, Accessible, Interoperable, Reusable)\cite{dempsey2022sharing} across the ML developmental lifecycle.
This paper builds on that foundation by introducing a data-centric framework that unifies datasets, features, workflows, models, and executions under a traceable, versioned system. Our goal is to enable lifecycle-aware reproducibility—not just for code re-runs, but for meaningful reconstruction of ML development across evolving teams and contexts.
Our key contributions include:
 \begin{itemize}
 \item An extensible data model structured around six core artifact types: Dataset, Feature, Workflow, Execution, Asset, and Controlled Vocabulary.
 \item Lifecycle-aware tracking and provenance mechanisms that preserve the relationships and evolution of artifacts.
 \item System implementation and evaluation through a clinical research use case.
 \end{itemize}
These contributions offer a structured, data-centric approach to reproducible machine learning in dynamic, collaborative scientific settings.

\section{Reproducibility in Scientific Machine Learning }
Machine learning (ML) is increasingly embedded in scientific research, making reproducibility a foundational principle for ensuring trust, accelerating discovery, and enabling collaboration. In eScience domains where decisions have high societal and economic impact, reproducibility is not only about verifying outcomes—it is essential for maintaining quality, transparency, and long-term sustainability of ML systems.

Despite this urgency, the reproducibility crisis persists~\cite{kapoor2023leakage}. In many cases, reproducibility is narrowly interpreted as the ability to re-run a script and obtain the same numerical results. This oversimplification leads to incomplete practices, where researchers may preserve data and code only for result regeneration. However, this approach often fails to support meaningful reuse, reinterpretation, or extension of the original work. Crucial details about how models were built, how data was prepared, and how decisions were made throughout the ML lifecycle are frequently missing. As a result, the knowledge embedded in the development process is lost, leaving others unable to follow or build upon the original path.

\begin{figure}[htbp]
    \centering
    \includegraphics[width=0.35\textwidth]{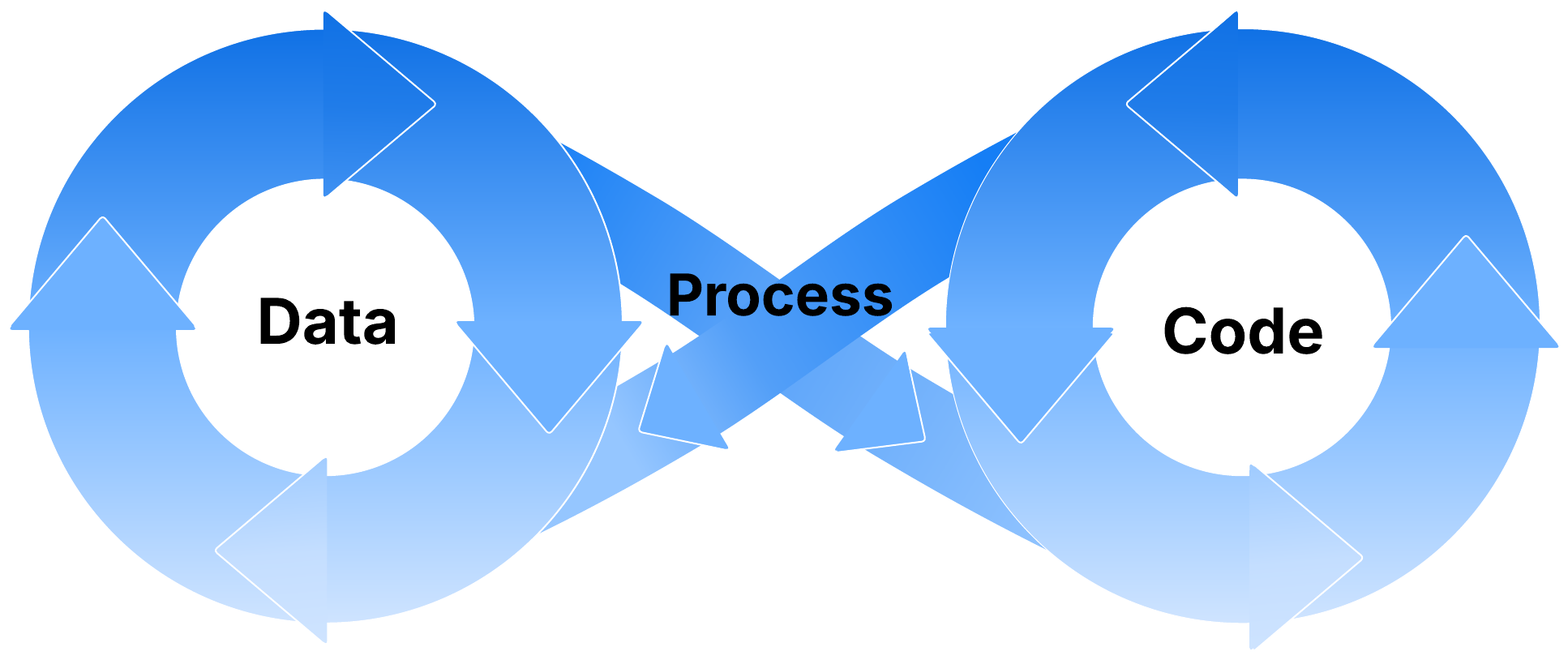}
    \caption{Conceptual model of reproducibility in machine learning, emphasizing the evolving interaction among data, code, and process. This dynamic relationship shows the iterative nature of scientific ML development and highlights the need for lifecycle-aware tracking.}
    \label{fig-repro-loop}
\end{figure}

To overcome this limitation, reproducibility in ML must be viewed more holistically, not as a static property of a single result, but as a dynamic process that captures how data, code, and human decisions interact over time. Unlike repeatability, which emphasizes consistency under identical conditions, or replication, which often involves different teams conducting similar experiments with new data, reproducibility is about tracing the original reasoning, design choices, and workflows. Reproducibility should be attained in multiple areas, including the outcome, analysis, and interpretation of results~\cite{gundersen2021fundamental}. In this view, reproducibility is not only about perfectly duplicating every computational step, but also about arriving at the same scientific conclusion through a documented and transparent path.
This perspective is illustrated in Figure~\ref{fig-repro-loop}, where reproducibility emerges from the interaction of data, code, and process. 
These elements—data, code, and process—do not operate in isolation. They evolve together as the scientific investigation unfolds, shaped by iterative experimentation, evolving goals, and interdisciplinary collaboration. Supporting reproducibility, therefore, goes beyond preserving static artifacts. It requires capturing the context in which decisions are made, results are interpreted, and knowledge is constructed. Rich metadata and structured records of development help ensure that these interactions remain transparent and traceable over time.
Understanding reproducibility requires examining the ML workflow itself, particularly how data preparation, modeling, and evaluation continuously shape the trajectory of scientific inquiry.

\section{ML Workflow Dynamics in eScience}
Scientific ML development is rarely linear or predictable. As discussed in the previous section, achieving reproducibility requires recording artifacts like data and code while also capturing the evolving and often iterative process through which ML experiments unfold. This is particularly critical in eScience domains, where domain experts and ML engineers collaborate to refine questions, engineer features, evaluate results, and revisit earlier decisions based on new findings. To address reproducibility in such dynamic environments, we examine the general structure of ML workflows and the challenges they present.

\subsection{Workflows in Collaborative ML}
\begin{figure*}[t]
    \centering
    \includegraphics[width=0.85\textwidth]{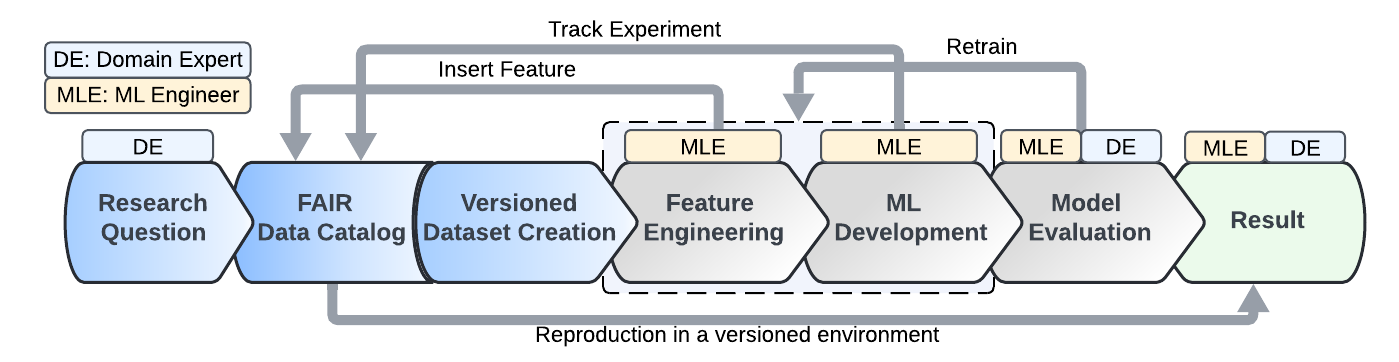}
    \caption{Machine learning lifecycle in collaborative eScience projects. The left-to-right flow represents the forward development process, where data is curated from the catalog and used to build models. The right-to-left arrows depict feedback loops, where evaluation outcomes inform upstream revisions to data, features, or modeling strategies. The grey box groups Feature Engineering and ML Development, reflecting their interdependence. Some feature extraction techniques require model development as part of the process. Together, these paths illustrate the bidirectional and iterative nature of ML workflows that complicates reproducibility.
}
    \label{fig-mllifecycle}
\end{figure*}
Figure~\ref{fig-mllifecycle} illustrates a representative ML workflow, starting from a research question and progressing toward model development, evaluation, and result reproduction. 
The left-to-right flow shows a typical development sequence: domain experts define the problem and upload initial datasets to a centralized FAIR catalog.
From there, ML engineers take over to perform dataset versioning, feature engineering, model training, and performance evaluation. Finally, the resulting models and insights are reviewed and potentially published or shared.

This development phase comprises several concrete workflows. Datasets are partitioned into training, validation, and testing subsets to prevent data leakage. 
Features are derived variables that capture structured information from raw data and serve as input to ML models.
Feature engineering transforms raw inputs into such variables, and ML models are trained using curated features, often accompanied by hyperparameter tuning and validation procedures. 
Evaluation follows, using standardized metrics to assess performance and reliability. Each of these stages involves specific roles—domain experts guide data interpretation and task framing, while ML engineers handle modeling, scripting, and system orchestration. This cross-functional structure creates a rich but complex pipeline of shared tasks and artifacts.

\subsection{Feedback Loops}
ML development does not stop at the first iteration. It is inherently bidirectional. As shown by the feedback arrows in Figure~\ref{fig-mllifecycle}, the right-to-left loops illustrate how new findings frequently lead to upstream revisions. These feedback loops are essential for iterative scientific reasoning, but they also introduce major challenges to reproducibility.

Three primary feedback loops complicate reproducibility by continuously reshaping ML experiments:
\begin{enumerate}
    \item \textbf{Insert Feature (\textit{Feature engineering to the FAIR data catalog})}: It is triggered when new features are derived or existing ones are redefined. This may occur in response to insights gained during exploratory analysis or as a refinement prompted by downstream model performance. These new features, once created, become part of the evolving dataset, which must be versioned to document their addition. Sending these changes back to the catalog ensures they are tracked, discoverable, and linked to their sources, preserving the provenance needed for interpretation and reuse.
    \item \textbf{Track Experiment (\textit{Model development to the FAIR data catalog})}: It typically arises at the conclusion of an experiment. Once a model has been trained, its outputs—such as learned weights, predictions, or performance metrics—along with the code, input data, and configuration that produced them, need to be registered. This step ensures that the experiment is reproducible. It is essential documentation for team hand-off, retrospective validation, and publication.
    \item \textbf{Retrain (\textit{Model evaluation to feature engineering and model development})}: It occurs when a model underperforms or behaves unexpectedly. In this case, teams often revisit prior steps to improve results. The feedback might lead to engineering better features, reconsidering label quality, filtering the data differently, or even choosing alternative modeling approaches. In such cases, the outputs of evaluation are not endpoints but prompts for a new round of experimentation. These iterations may reuse the same workflow while operating on different configurations or data versions.
\end{enumerate}

These forward and feedback paths form a bidirectional, collaborative development system. Artifacts are refined and reused across cycles, and they are also interdependent: changes in one stage ripple through the rest. Without consistent and structured tracking, these dependencies quickly become opaque. Missing context from one experiment can disrupt downstream development and obscure the rationale behind decisions. The cumulative effect creates a reproducibility bottleneck that grows with each iteration.

These challenges highlight the necessity of strong orchestration of the development cycles, especially the input and output artifacts connected by workflow loops. Data, features, and code are not static inputs or outputs—they evolve as part of the process and define each experiment. Capturing them accurately is essential not just for re-running workflows, but for understanding and reproducing the scientific reasoning they encode. 
Lifecycle-aware reproducibility requires shifting from workflow-centric orchestration to structured, traceable data artifacts.
This motivates the data-centric approach introduced in the next section, where artifacts become the foundation for transparency, reusability, and accountability in ML-driven science.

\section{A Data-Centric Reproducible ML Framework\label{sec:data_centric}}
\subsection{Principles of the Data-Centric Approach}
The complexity of ML workflows, driven by iterative development and collaborative refinement, calls for a reproducibility strategy that goes beyond static artifacts. Rather than treating reproducibility as a matter of re-executing code, we approach it as the ability to systematically manage the evolving relationships between data, code, and process. This motivates a data-centric approach to ML development.

In model-centric or process-centric infrastructure, data is often treated as static input and results as terminal outputs. Workflow scripts and modeling logic revolve around these endpoints, with limited attention to how artifacts evolve. 
However, in practice, data artifacts are continuously transformed and reused as experiments progress.
As an alternative, we propose a data-centric model in which experimental data, such as datasets, features, models, and outputs, are the primary artifacts being managed. In a data-centric approach, every data artifact is an identifiable, versioned, arbitrary collection of data elements and associated metadata. Interfaces and services are defined to enable both human and computational agents to curate, identify, evolve, consume, and produce datasets.

A data-centric approach enables detailed provenance tracking, clearer collaboration, and more reliable reproducibility. Rather than viewing the ML lifecycle as a sequence of isolated steps, it frames development as a network of evolving, interrelated data artifacts. By explicitly structuring these artifacts and their connections, the framework provides a standard structure that can unify all workflows, forward and feedback within a single conceptual model.

\subsection{Structuring ML Workflows Through a Data Model}

A key element of our data-centric approach is an explicit data model that captures core artifact types and their interrelationships. This model serves as the foundation for implementing lifecycle-aware reproducibility, enabling consistent tracking, versioning, and reuse of artifacts throughout iterative ML development. We define the data model and provide mechanisms so that it can evolve over time as the understanding of the ML model requirements evolves. 

\begin{figure}[htbp]
    \centering
    \includegraphics[width=0.47\textwidth]{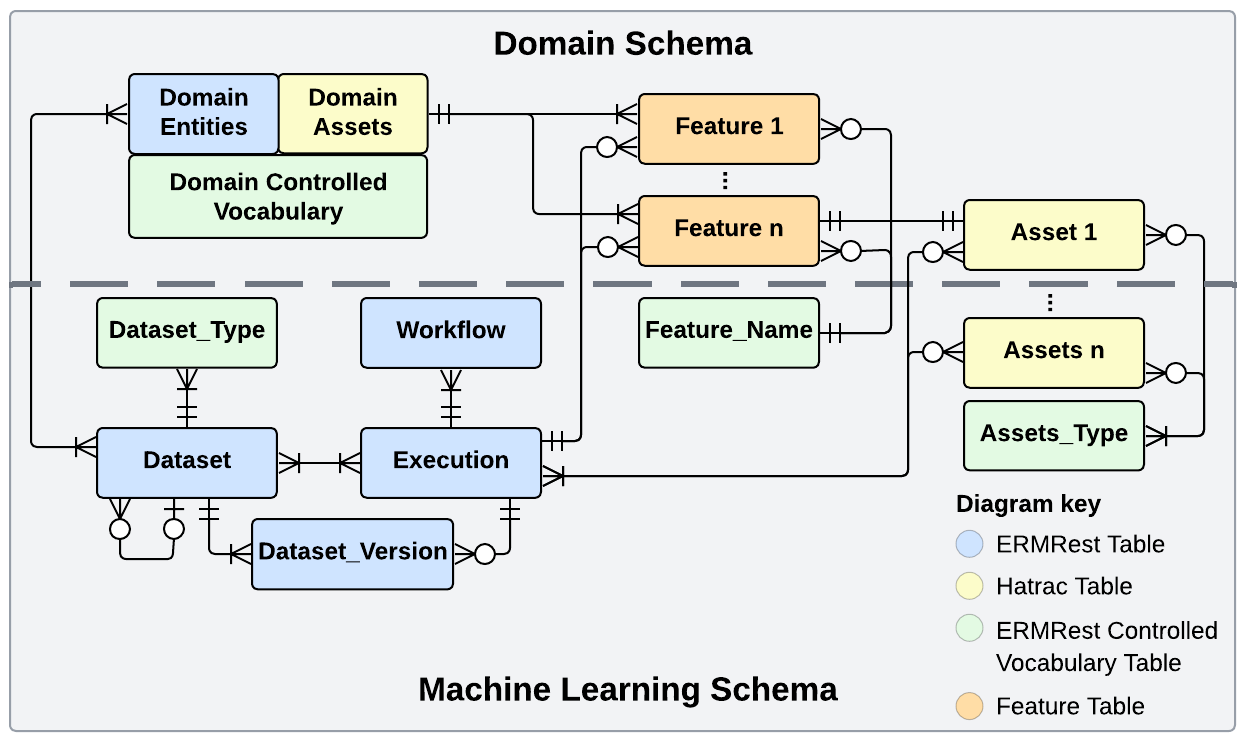}
    \caption{High-level data model supporting lifecycle-aware reproducibility through structured artifacts. The model is divided into two interlinked schemas: a domain schema for application-specific entities and an ML schema for standardized ML artifacts. The six core artifact types—Dataset, Feature, Workflow, Execution, Asset, and Controlled Vocabulary—enable versioning, provenance tracking, and reproducible reuse across iterative ML development.}
    \label{fig-ERD}
\end{figure}

Figure~\ref{fig-ERD} outlines a high-level data model to support our data-centric architecture for reproducible ML. This model explicitly separates domain-specific entities from generalized ML workflow artifacts, organized into two complementary layers:
\begin{enumerate}
    \item \textbf{Domain Schema}: Captures entities unique to specific application domains, allowing for customization tailored to the needs of particular scientific or analytical contexts, without sacrificing overall schema consistency.
    \item \textbf{ML Schema}: Represents standardized ML workflow artifacts common across various applications, promoting consistent reuse and ensuring reproducibility through structured tracking of ML processes.
\end{enumerate}
These schemas interlink through clearly defined relationships, forming a robust yet adaptable foundation for managing diverse ML lifecycles and workflows across domain applications. 
To maintain both flexibility and robustness in our data-centric design, we represent core activities within the ML lifecycle through six primary artifact types:
\begin{itemize}
    \item \textbf{Dataset}: A curated collection of data used in ML tasks, supporting versioning and hierarchical structure for accurate usage and reuse.
    \item \textbf{Feature}: A derived variable created from source data, serving as ML model input and often evolving over iterations.
    \item \textbf{Asset}: A non-tabular file or output (e.g., trained models, predictions, images) linked to other artifacts for traceability of unstructured data.
    \item \textbf{Workflow}: A formalized sequence of executable steps (e.g., a notebook or script) that defines an ML experiment or pipeline.
    \item \textbf{Execution}: A recorded instance of a workflow run, capturing its inputs, outputs, configuration, and context.
    \item \textbf{Controlled Vocabulary}: A shared terminology for labeling and categorizing artifacts.
\end{itemize}

As illustrated in Section 2 (Figure \ref{fig-repro-loop}), reproducibility depends not only on preserving artifacts individually but also on capturing how they interact and evolve together. The data model in Figure~\ref{fig-ERD} reflects this principle by explicitly explaining relationships between datasets, code, features, and outputs. These connections support structured versioning, consistent metadata, and full traceability across the ML lifecycle, ensuring reproducibility that extends beyond static records. To support this interaction-driven perspective, the artifacts are not isolated in the data model but are part of interdependent elements. Their relationships describe how they are created, modified, and reused throughout the ML process. Executions record how datasets, workflows, features, and assets are used in a specific run, and what new version of the datasets, features, and assets are generated. Its primary role is documenting how artifacts come together, enabling clear traceability of each experimental step and outcome.

This relational structure captures the iterative nature of ML development and the feedback loops that occur as experiments evolve. Every change can be traced, whether in data partitioning, feature design, or modeling approach. This makes it possible to reconstruct not just the outputs of an experiment but also the reasoning and decisions behind them. In this way, the framework supports both forward development and backward analysis, creating a foundation for transparent and reproducible ML research.


\section{Platform Foundations: Deriva-ML}

In response to the challenges, the initial Deriva-ML platform was developed as a socio-technical ecosystem for ML that supports collaborative workflows and ensures that all the data adhere to the FAIR principles—Findable, Accessible, Interoperable, and Reusable~\cite{wilkinson2016fair}. As illustrated in~\cite{li2024deriva}, the architecture integrated key services such as ERMRest (a metadata catalog service), Hatrac (a versioned object store), and Chaise (a model-driven user interface), along with the Deriva-ML Python library and external ML computing environments. This architecture enabled structured metadata annotation, discoverability, and curated data exchange within a collaborative research setting.

The Python-based Deriva-ML library provided high-level abstractions for data access and model development. It allowed ML researchers to interact with the platform using familiar tools and scripts while ensuring their activities were tracked and reproducible. Role-based access control and data curation workflows further supported coordination across diverse contributors, including domain experts, ML engineers, and research software engineers.

The platform achieved several important milestones:
\begin{itemize}
    \item Continuous FAIRness: From datasets to execution outputs, were structured and versioned with persistent identifiers and standardized metadata.
    \item Socio-technical Collaboration: The system supported differentiated roles and workflows to bridge the gap between domain experts and engineers.
    \item Flexible Data Access: The platform enabled users to interact through both web interfaces and programmatic APIs.
\end{itemize}

Despite these strengths, the system exhibited limitations when applied to the full ML lifecycle. The original platform lacked mechanisms to fully trace the lineage of evolving datasets, features, and workflows across iterations. It also provided limited support for managing the Workflows and Executions on the compute platform. Those limitations hindered reproducibility in dynamic ML workflows.

To address these limitations, the enhanced platform captures the linkage between datasets, features, and code and supports detailed provenance, configurable workflows, and robust versioning of all ML artifacts. 
This design enables reproducible, iterative experimentation and sets the stage for the following sections.

\section{Implementation: Artifact Versioning and Provenance}

Section~\ref{sec:data_centric} identifies the core abstractions that we believe to be important for a data-centric ML environment: datasets, features, workflows, and executions. This section describes the specific mechanisms used to define, version, and trace these artifacts throughout the ML lifecycle. We begin by detailing how datasets and features are stored and versioned, and then describe how workflows and executions are orchestrated to create fully traceable ML pipelines.

\subsection{Lifecycle Management of Dataset}

\subsubsection{Structure}
In Deriva-ML, a Dataset is defined as a collection of data elements. These collections can be multimodal, encompassing diverse data elements from different data tables or across different granularities (e.g., a collection of subjects as well as a collection of images). Every dataset is automatically assigned a resolvable, globally unique identifier.
A dataset includes not only the assets of interest but also all of the relationships and metadata associated with those assets.
The system allows a dataset to contain references to other datasets, forming a nested structure that allows the construction of parent datasets that logically group multiple child datasets. This approach facilitates a more precise representation of the coverage of the Dataset, simplifies data management, and enhances the precision with which subsets are assigned to training, validation, or evaluation tasks. 

\subsubsection{Dataset Versioning}
Datasets evolve over time due to updated relationships, metadata, or new data acquisitions. Deriva-ML employs a semantic versioning strategy to track and manage changes~\cite{semanticversioning} to datasets in which dataset versions are identified via a triple: \texttt{(major, minor, patch)}. Managing versions in Deriva-ML is complicated due to two factors: 1) a dataset may be included in more than one nested Dataset, and 2) the version must capture all of the relationships and metadata as they existed when the version was created. The first issue is addressed by implicitly incrementing the version of all of the sibling and parent datasets that may be connected to the Dataset being versioned. The second is addressed by the snapshot capability in Deriva~\cite{czajkowski2018ermrest}, which allows Deriva-ML to efficiently capture the complete state of a Deriva catalog at version creation time, ensuring that historical configurations remain accessible for auditing and reproduction. 
Each dataset version links to the Execution that created it, ensuring explicit traceability to the computational process that generated it.
A dataset history table in the catalog records all of the dataset versions, catalog snapshot identifiers, and execution IDs for every Dataset in the catalog.

\subsubsection{Caching}
Typical ML workflows require dedicated computational resources and may consider multiple computations across the same data, making fetching potentially large datasets from the Deriva-ML catalog repeatedly impractical. To support consistent and efficient reuse of versioned datasets across computing environments, Deriva-ML provides a simple caching mechanism using a checksum of an entire dataset version.  
Checksums are calculated from the dataset metadata and a manifest of versioned, checksummed assets using the BagIt packaging format~\cite{chard2016ll} and BDBag tools.
Additionally, we utilize Minimal Identifiers (Minids) as lightweight, permanent, resolvable identifiers. We also record the location, checksum, and length of each generated BDBag.

The caching mechanism follows a two-stage resolution process. When a user references a dataset using its Derivam-ML record ID and version number, the system first checks whether a Minid is already registered for that version. If a Minid exists, the checksum is retrieved from the Minid record and compared to the checksums in a specified cache directory. If no match is found, the Dataset is copied by resolving the Minid and retrieving the associated BDBag directly and \textit{materializing} it by retrieving all of the assets in the manifest. This approach ensures the user receives a validated and consistent copy of the Dataset without the hassle of preparing a new bag.
If no Minid is found, Deriva-ML initiates the bag preparation process. The newly generated Minid will be updated to the Version table. The bag is then materialized based on the newly registered Minid. This strategy allows any subsequent user, regardless of whether they share the same cache directory, to bypass bag preparation and retrieve the exact dataset contents using the Minid, thereby improving efficiency and consistency across teams.

The metadata and relationships for a materialized bags on a compute-local cache are loaded into a lightweight SQLite database to support fast local access. Using the Deriva-ML datasetbag interface, the data table can be easily read into Python Pandas dataframes or TensorFlow datasets. This design enables seamless integration into ML workflows without requiring repeated file parsing or manual data wrangling.
This implementation ensures that every versioned Dataset is unambiguously identifiable, reproducibly packaged, and efficiently accessible, whether reused by the same user or across collaborators, while preserving the fidelity and provenance of data across execution environments.

Through this implementation, Deriva-ML ensures that all datasets are explicitly versioned, reproducible, and interoperable across the ML lifecycle. The system's ability to track version history and support portable caching significantly enhances the transparency and reliability of ML experiments, particularly in collaborative settings involving complex, evolving datasets.

\subsection{Feature}
\begin{figure}[htbp]
    \centering
    \includegraphics[width=0.45\textwidth]{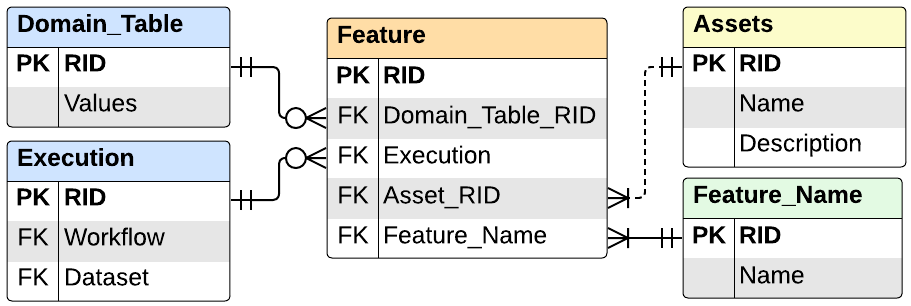}
    \caption{Feature data model in Deriva-ML. Each Feature is defined by four key connections: (1) a unique feature name, (2) a reference to the source record in a domain-specific table, (3) the Execution that generated it, and (4) the feature value, which may be a scalar, a controlled vocabulary term, or a file-based object represented by an Asset entity.}
    \label{fig-featureERD}
\end{figure}
Features are structured artifacts representing derived or computed values associated with domain data. Each Feature captures the full context of its origin and usage, ensuring its interpretability, traceability, and reusability across experiments. As illustrated in Figure~\ref{fig-featureERD}, each Feature is defined by four connected components: (1) the Feature's name, (2) the source domain table, (3) the Execution that generated it, and (4) the Feature's values, which may consist of one or more controlled vocabulary terms, file based objects or assets, or numerical or text fields. 
An API call in the Deriva-ML library allows for the dynamic creation of new features.

This flexible design supports multiple derived features from the same source field, accommodating iterative feature engineering, comparative evaluation of feature sets, and enrichment through manual labeling to improve data quality. 
Each feature instance is tied to a specific source record in a domain table (e.g., a raw clinical image from which distinct pathological regions are cropped and stored as individual features).
The association with the Execution table preserves a complete provenance trail, capturing the workflow, parameters, and data used to generate the feature. 
Additionally, the use of controlled vocabularies standardizes terminology across sources and contributors, enhancing the interoperability of features across workflows and teams.


\subsection{Workflow and Execution}
\subsubsection{Workflow}
In Deriva-ML, a workflow represents a reusable and version-controlled computational procedure, typically implemented as a script or notebook, or pipeline. Each entry in the Workflow table includes a URL pointing to the source code (e.g., a GitHub repository), along with a checksum and version label to ensure code integrity and reproducibility.
Workflow versioning can be managed manually, for example, through Git-based version control and updates to the catalog, or automatically recorded during Execution to reflect the specific version used in a given run.

\subsubsection{Execution}
An execution represents a specific run of a workflow and captures the complete context in which it was applied. Each Execution is directly linked to a workflow and records all associated inputs and outputs. To accurately track the data and asset dependencies for a given run, executions are initiated using a structured configuration file. This configuration includes references to the dataset record ID and version, any relevant input assets, and the corresponding workflow object.

Once initiated, Deriva-ML resolves the specified inputs, retrieves versioned datasets and assets, and stages them into the compute environment. The workflow is then executed, typically involving model training, feature extraction, or evaluation, and upon completion, the resulting output files are uploaded back to the catalog. Along with the outputs, the current version and checksum of the workflow script or notebook are saved to ensure that any changes made during development are recorded and linked to the Execution. 

While commonly used in ML development, unstructured use of notebooks can significantly impair reproducibility. Rather than creating an overly constrained environment, Deriva-ML provides a simple combination of mechanisms and best practices to enhance notebook reproducibility. The Deriva-ML distribution includes the tool \texttt{papermill}~\cite{papermill}, which can inject specified parameters into a notebook and run the notebook from the first cell to the last. Users are encouraged to commit notebooks prior to running, and Deriva-ML will automatically strip output values from notebook cells, and use papermill to execute notebooks as scripts, allowing users to develop in tools that they are familiar with, but providing some degree of reproducibility as long as a few simple rules are followed.


\subsection{Asset Handling and Integration}
Assets in Deriva-ML represent digital files generated or consumed during workflow executions, and each asset is explicitly linked to the Execution that defines its provenance and context of use. Assets may optionally be connected to features when the file itself constitutes a feature value, such as a structured annotation or visual artifact.

The asset framework is designed to support arbitrary file formats, allowing flexible integration of model artifacts, configuration files, intermediate results, or external outputs. 
Each asset is associated with a role, either "input" or "output", which is recorded in its linkage to the corresponding Execution. This role specification enables precise tracking of how each file was used in the ML pipeline. 
For instance, consider a scenario where a convolutional neural network (e.g., a VGG19 model saved as an MD5-named file) is trained during an ML experiment. The resulting model file is stored as an asset in a designated model Asset table and linked as an output of the training Execution. It can be reused in subsequent prediction executions. 
Asset downloads are coordinated using record IDs specified in execution configuration files. Uploading assets is managed via the Hatrac storage system. 
Each asset table is mapped to a designated directory in the compute environment, and the Deriva-ML API automates the process of identifying and uploading the relevant files. Once Execution is complete, all files in the corresponding directory are transferred to the catalog, where they are stored alongside metadata and linked to the generating Execution. This design ensures that asset provenance, usage roles, and file-level metadata are systematically recorded, enabling reproducible workflows and effective reuse of artifacts across iterative experiments.

\subsection{Controlled Vocabulary}
Controlled vocabularies (CVs) are used throughout the ML and domain schema to standardize terminology and ensure consistency across artifacts. 
They are applied to categorize dataset metadata, annotate Feature values, label Assets, and describe Workflow and Execution. 
Deriva provides a UI and a programmatic API for managing vocabulary, supporting operations such as creating new vocabulary, listing all existing CVs, and adding or querying terms.
By enforcing consistent labeling, controlled vocabularies facilitate accurate classification of artifacts and improve human and machine interpretability. Implementing the Controlled Vocabulary streamlines the collaborative work and enables more effective discovery by making artifacts more FAIR.



\section{Use Case}
\subsection{EyeAI: AI-Powered Glaucoma Detection}
We evaluate Deriva-ML as the core infrastructure for the EyeAI project at the Keck School of Medicine of the University of Southern California (USC), aiming to develop scalable AI models for glaucoma detection. The project brought together ophthalmologists, ML engineers, and data scientists to collaboratively build and evaluate diagnostic models using clinical data from Los Angeles County and the Keck School of Medicine at USC.

Two core challenges defined the project. First, glaucoma diagnosis relies on multimodal and noisy data, including fundus photographs, clinical tests, and subjective assessments. Deriva-ML addressed this by providing a unified domain data model to manage cross-modality integration, versioned datasets, and consistent data organization. Second, the interdisciplinary nature of the team required seamless collaboration and artifact reuse. 

The primary ML task was to assess the performance and label efficiency of two vision models—VGG19 and RETFound~\cite{zhou2023foundation}. It followed a two-stage pipeline: (1) extract the optic nerve using an object detection model applied to fundus images, and (2) classify referable glaucoma using the cropped optic nerve regions. These tasks are detailed in our prior study on deep learning and clinician performance in glaucoma detection~\cite{nguyen2025comparison}. Datasets were organized in a parent-child structure with globally unique identifiers for training, validation, test, and ad-hoc subsets, preventing data leakage and enabling reproducible reuse throughout the iterative development process.

Major experiments were conducted using Deriva-ML's structured artifact tracking:
\begin{itemize}
    \item Labeling and Validation: Thirteen clinicians independently graded the test dataset using EyeAI's reviewing tool. Labels were stored as Image Diagnosis Features, enabling gold-standard construction, inter-rater comparison, and full provenance.
    \item Feature Engineering – Optic Nerve Detection: A detection model was trained to localize optic nerve regions. Bounding boxes were stored as Image Annotation Features linked to SVG assets and the Execution that generated them, supporting storage, inspection, traceability, and iterative refinement.
    \item Model Training and Evaluation (VGG19 and RETFound): Structured workflows for both architectures allowed configurable re-execution, enabling model comparison and performance evaluation.
    \item Label Efficiency and Robustness Testing: To study the impact of data volume, 142 models were trained across 36 subsets; 71 models were re-trained for RETFound by modifying only execution configurations.
    \item Debugging and Feature Re-Extraction: When performance degraded, execution records traced the issue that stemmed from feature extraction. The detection model was reparameterized and re-trained, and downstream experiments were repeated with complete consistency and minimal effort.
\end{itemize}

This use case highlights how Deriva-ML supports complex ML workflows in real-world research. It enabled scalable experimentation, transparent collaboration, and reproducible debugging across a multidisciplinary team. Key contributions included reusable feature artifacts (Image Diagnosis and Image Annotation), configurable and standardized training workflows, and full execution provenance.

The EyeAI project produced 70+ datasets, 130 executions, and 142 trained models—all versioned, traceable, and accessible via the Deriva-ML catalog. What could have been a fragmented and error-prone development process was transformed into a cohesive, auditable, and collaborative workflow, validating the role of Deriva-ML as a foundational tool for reproducible ML research. 
\subsection{Optic Nerve Feature Extraction}
\begin{figure*}[t]
    \centering
    \includegraphics[width=1\textwidth]{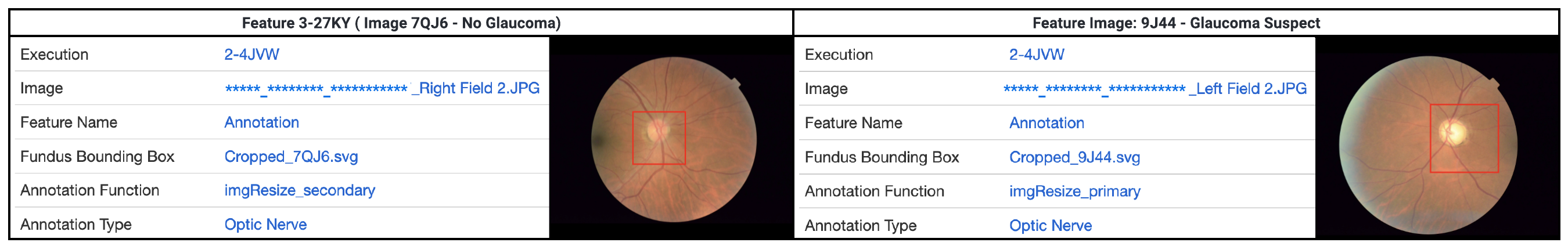}
    \caption{The Feature record detail page in the Deriva Chaise interface has a visualization of the optic nerve bounding box annotated on a fundus photograph. This interface presents the complete metadata of a Feature, including its generating Execution, source domain entity (the fundus image used for optic nerve detection), Feature Name (Annotation), Feature Value (an Asset), and links to two Controlled Vocabularies—Annotation Function (indicating the cropping method) and Annotation Type (optic nerve). The page ensures end-to-end transparency and reproducibility of feature generation in ML workflows.}
    \label{fig-opt-nerve}
\end{figure*}
To illustrate a specific ML development process within the EyeAI project, we detail the optic nerve feature extraction process.
To precisely identify the optic nerve region in fundus photographs, we developed a template-based cropping method to estimate the optic nerve center. An initial candidate set of bounding boxes is generated around the estimated optic nerve center using sliding window techniques. 
Each candidate was then scored by a pre-trained VGG19 model that evaluates crop quality based on a human-annotated training set. 
The VGG19 crop evaluation model was trained using a curated dataset of high-quality optic nerve crops. This Dataset was uploaded to the EyeAI catalog and used to launch a training workflow via Deriva-ML. All key metadata—including the dataset version, model parameters, and input files—were automatically linked to the corresponding execution record. Upon completion, the trained model checkpoint was uploaded to the catalog as an Execution Asset, and the Workflow notebook was versioned and pushed to a Git repository. This ensured that both the model object and its workflow logic were findable, reusable, and reproducible across future tasks.

The trained VGG19 model was used within a Deriva-ML prediction Workflow to extract optic nerve bounding boxes from fundus images for deployment. Each detected region was stored in the catalog as an Image Annotation Feature, linked to its corresponding image and execution. This feature captured both the semantic meaning of the annotation and its provenance, including the algorithm and parameter settings used, as shown in Figure~\ref{fig-opt-nerve}.
Deriva-ML ensured all outputs were versioned and traceable, enabling visual inspection, metadata enrichment, and structured reuse across downstream workflows.

After initial deployment, we observed that the diagnosis model underperformed on a new dataset. Using the saved artifacts in Deriva-ML, we traced the issue back to the upstream optic nerve extraction step. Though effective on the training set, the bounding box model failed to generalize. This prompted a refinement cycle where the detection model was re-trained, and downstream executions were re-run, demonstrating the reproducibility and adaptability enabled by the system.

\section{Related Work}

Many existing reproducibility tools are model- or process-centric, focusing on tracking trained models, code artifacts, and execution configurations. These tools aim to streamline experimentation and support deployment pipelines, often through lightweight APIs and intuitive dashboards. Platforms such as MLflow~\cite{zaharia2018accelerating}, Weights \& Biases~\cite{biewald2020wandb}, and Comet~\cite{cometml2024} enable users to log hyperparameters, evaluation metrics, and model versions to facilitate experiment comparison and iteration. However, they generally assume fixed data schemas and provide limited mechanisms for managing evolving datasets, fine-grained data provenance, or schema-level metadata. As a result, they may fail to capture upstream sources of variation or error, including data leakage, inconsistent label definitions, or shifting cohort boundaries~\cite{sculley2015hidden}.

In contrast, data-centric tools position the data, not the model, as the primary artifact of machine learning development. Systems like DVC~\cite{barrak2021co}, OpenML~\cite{vanschoren2014openml}, and CodaLab~\cite{pavao2023codalab} emphasize robust Dataset versioning, structured metadata capture, and reproducible workflows that document the full data transformation pipeline. These platforms facilitate the inspection, reuse, and comparison of datasets across experiments, improving transparency in how models are trained and evaluated. Recent work in data-centric AI~\cite{jakubik2024data,jarrahi2023principles} has highlighted the importance of systematically managing data quality, lineage, and evolution as key factors in building reliable and reproducible ML systems.

Despite these advances, current tools often operate in isolation, requiring researchers to integrate multiple systems for managing data, code, and environments. Reproducibility is increasingly recognized as a socio-technical challenge~\cite{sculley2015hidden}, requiring better tooling, structured workflows, evolving metadata, and collaborative coordination.

\section{Conclusion and Future Work}
This paper introduces a data-centric framework for managing the entire machine learning lifecycle in collaborative eScience, with a focus on enabling reproducible feedback loops across data, features, workflows, and model executions. 
Using a structured model of six key artifact types—Dataset, Feature, Workflow, Execution, Asset, and Controlled Vocabulary—we ensure complete traceability and versioning across iterative development. 
Our implementation in real-world clinical use cases demonstrates improved transparency, coordination, and reproducibility.

Reproducing ML environments (e.g., TensorFlow) remains challenging without deeper support for containerized execution. Execution configuration and metadata linkage still require manual intervention, and validation of data usage is also performed manually. Future work will focus on environment packaging, further automating configuration executions, and building tools to visualize artifact lineage. Broader evaluations across domains are planned to assess usability, extensibility, and integration into diverse scientific applications.


\section*{Acknowledgment}

The authors wish to thank the members of the EyeAI research team who assisted in the development of the evaluation use case.

\bibliographystyle{IEEEtran}

\end{document}